\documentclass[10pt,twocolumn,letterpaper]{article}

\usepackage{cvpr}              

\usepackage{graphicx}
\usepackage{amsmath}
\usepackage{amssymb}
\usepackage{booktabs}
\usepackage{multirow}
\usepackage{makecell}
\usepackage{colortbl}
\usepackage{color}
\usepackage[accsupp]{axessibility}
\definecolor{mygray}{gray}{.93}
\usepackage[pagebackref,breaklinks,colorlinks]{hyperref}

\newcommand*\samethanks[1][\value{footnote}]{\footnotemark[#1]}
\author{Foo\thanks{University of Podunk, Timbuktoo}
	\and Bar\samethanks
	\and Baz\thanks{Somewhere Else}
	\and Bof\samethanks[1]
	\and Hmm\samethanks}

\usepackage[capitalize]{cleveref}
\crefname{section}{Sec.}{Secs.}
\Crefname{section}{Section}{Sections}
\Crefname{table}{Table}{Tables}
\crefname{table}{Tab.}{Tabs.}

\def\cvprPaperID{4067} 
\def\confName{CVPR}
\def\confYear{2022}

\usepackage{color}

\begin{document}
	
	\title{Shunted Self-Attention via Multi-Scale Token Aggregation}
	
	\author{Sucheng Ren$^{1,2*}$,~~~Daquan Zhou$^1$\thanks{The first two authors contributed equally.},~~~Shengfeng He$^2$,~~~Jiashi Feng$^{3}$\thanks{Corresponding author.},~~~Xinchao Wang$^1$\samethanks \\
		$^1$National University of Singapore, 
		$^2$South China University of Technology,
		$^3$ByteDance Inc. \\
		{\tt\small 
			oliverrensu@gmail.com,
			daquan.zhou@u.nus.edu,
			shengfenghe7@gmail.com,}\\
		{\tt\small jshfeng@gmail.com,
			xinchao@nus.edu.sg}}
	\maketitle
	
	\begin{abstract}
		
		Recent Vision Transformer~(ViT) models have demonstrated 
		encouraging results across various computer vision tasks,
		thanks to its competence in modeling long-range dependencies 
		of image patches or tokens via self-attention. 
		These models, however, usually designate the similar receptive fields of each token feature within each layer.
		Such a constraint inevitably 
		limits  the ability of each self-attention
		layer in capturing multi-scale features, 
		thereby leading to
		performance degradation in handling images 
		with multiple objects of different scales. To address this issue, 
		we propose a novel and generic strategy, termed shunted self-attention~(SSA), that allows ViTs to  model the attentions at  hybrid scales per attention layer. The key idea of SSA is to inject heterogeneous receptive field sizes into tokens: before computing  the self-attention matrix, it selectively merges tokens to represent larger object features while  keeping certain tokens to preserve fine-grained features. This novel merging scheme enables the self-attention   to learn relationships between objects with different sizes, and simultaneously reduces the token numbers and the computational cost. Extensive experiments across various tasks demonstrate the superiority of SSA. Specifically, the SSA-based transformer achieve 84.0\% Top-1 accuracy and outperforms the state-of-the-art Focal Transformer on ImageNet with only half of the model size and computation cost, and surpasses  Focal Transformer by 1.3 mAP on COCO and 2.9 mIOU on ADE20K under similar parameter and computation cost. Code has been released at \href{https://github.com/OliverRensu/Shunted-Transformer}{https://github.com/OliverRensu/Shunted-Transformer}.

	\end{abstract}
	
	\section{Introduction}
\label{sec:intro}
\begin{figure}[t]
    \centering
    \begin{tabular}{c}
        \includegraphics[width=0.48\textwidth]{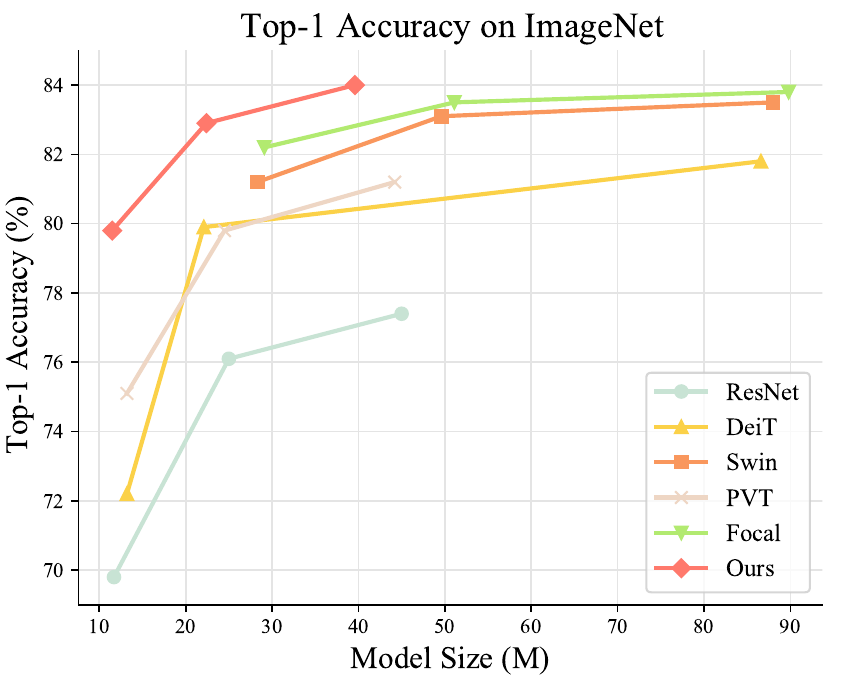}
    \end{tabular}
    \caption{Top-1 accuracy on ImageNet of recent SOTA CNN and transformer models. Our proposed Shunted Transformer   outperforms all the baselines including the recent SOTA   focal transformer (base size). Notably, it achieves competitive accuracy to DeiT-S with 2$\times$ smaller model size.}
  \label{fig:sota}
  \vspace{-2mm}
\end{figure}

The recent Vision Transformer (ViT) models~\cite{vit} 
have demonstrated   superior performance across various computer vision tasks, \eg,  
classification~\cite{deng2009imagenet}, object detection~\cite{coco,pascal}, semantic segmentation~\cite{ade20k,cityscapes} 
and video action recognition~\cite{action1,action2}. Different from convolutional neural networks focusing on local modeling, ViTs partition the input image into a sequence of patches (tokens) and progressively update the token features via global self-attention. The self-attention  can     effectively model  long-range dependencies of the   tokens and progressively expand sizes of  their receptive fields    via aggregating information from other tokens, which accounts largely for the success of ViTs.

However, the self-attention mechanism 
also brings the cost  of expensive  
memory consumption that is quadratic     
w.r.t.\ the number of input tokens. 
Thus, state-of-the-art Transformer models have  
resorted to various down-sampling strategies  to reduce
the  feature size and the memory consumption.
For example, the approach of~\cite{vit} 
conducts a 16$\times$16 down-sampling
projection at the first layer,
and computes the self-attention at  the resulted 
coarse-grained and single-scale
feature maps;  the incurred feature information loss,
therefore, inevitably downgrades 
the model performance.  Other approaches strive  
to compute self-attention at high-resolution features 
and reduce the cost by merging tokens with spatial 
reduction on tokens~\cite{pvt, pvtv2,focal}. 
Nevertheless, these approaches tend to merge 
too many tokens within one self-attention layer, 
thereby resulting in a mixture of 
tokens from small objects and background noise. 
Such behavior, in turn, makes
the model less effective in capturing  small objects.

Besides, prior Transformer models
have largely overlooked the 
multi-scale nature
of scene objects
within on attention layer,
making them 
brittle to in-the-wild
scenarios that involves 
objects of distinct sizes.
Such incompetence is, technically,
attributed to 
the their underlying attention mechanism:
existing methods rely on only
\textit{static} receptive fields of the tokens
and uniform information granularity
within one attention layer,
and are therefore 
incapable of capturing 
features at different scales simultaneously.  

\begin{figure}[t]
    \centering
    \begin{tabular}{ccc}
        \hspace{-0.2mm}\includegraphics[width=0.15\textwidth]{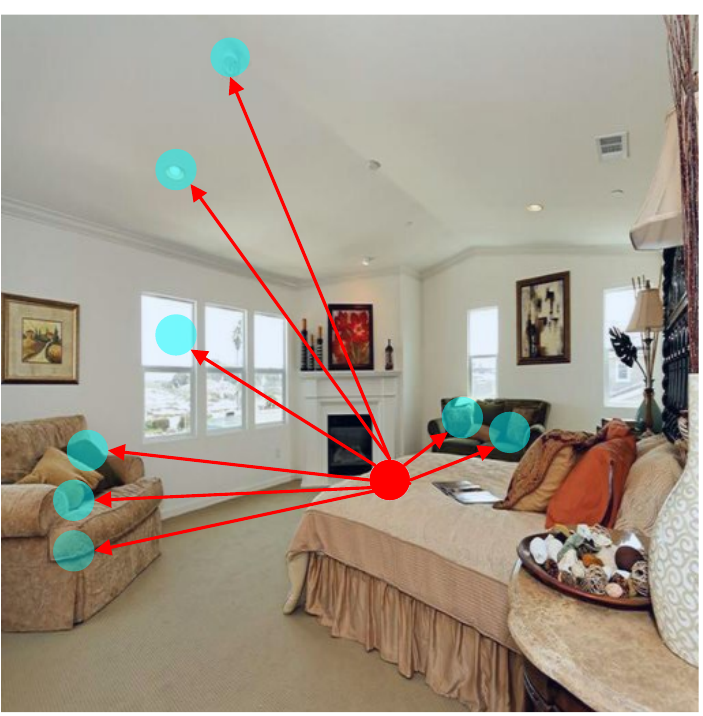}&\hspace{-4.2mm}
        \includegraphics[width=0.15\textwidth]{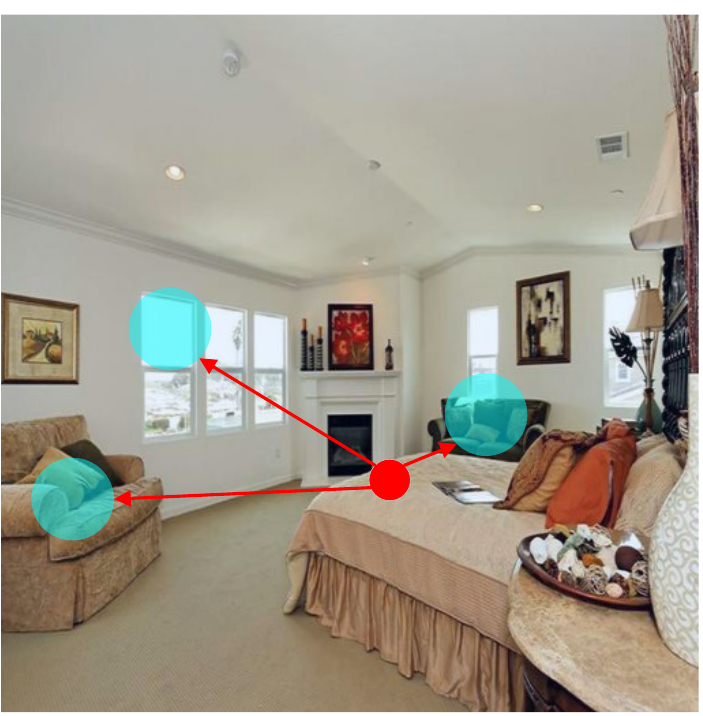}
        &\hspace{-4.2mm} \includegraphics[width=0.15\textwidth]{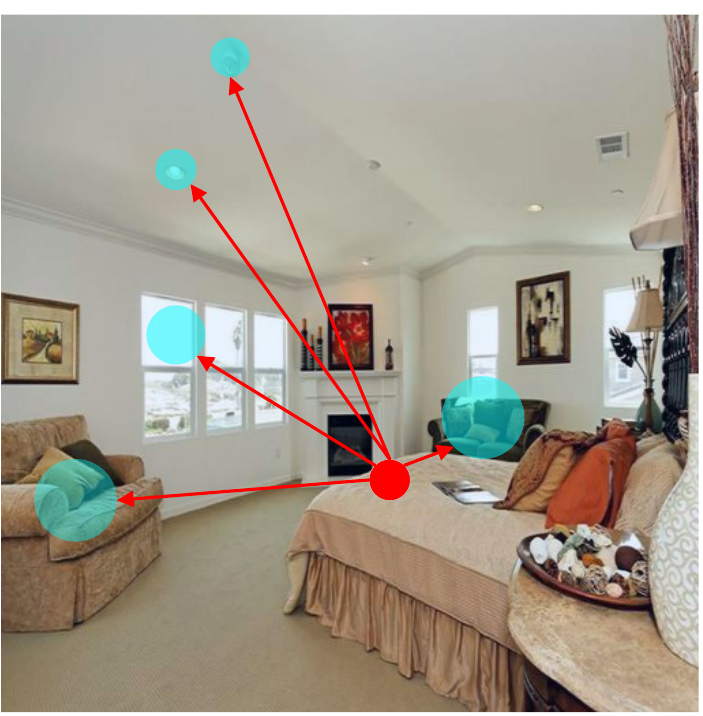}\\
        \hspace{-0.2mm}(a) ViT &\hspace{-4.2mm} (b) PVT &\hspace{-4.2mm} (c) Ours
    \end{tabular}
    \caption{Comparison of different attention mechanisms in Vision Transformer (ViT), Pyramid Vision Transformer (PVT), and our SSA with the same feature map size. The number of circles represents the number of tokens involved in the self-attention computation, and reflects  the computation cost. The size of the circle indicates the receptive field size of the corresponding token. Unlike ViT and PVT, our method adaptively merges circles on large objects for enhancing computation efficiency, and accounts for objects of different scales simultaneously.}
  \label{fig:comparison}
\end{figure}

To address this limitation, we introduce a novel and generic 
self-attention scheme, termed shunted self-attention~(SSA),
which explicitly allows the self-attention heads 
within the same layer to respectively  
account for  coarse-grained and
fine-grained features.
Unlike
prior methods that merge too many tokens or 
fail in capturing small objects,
SSA  effectively
models objects of various scales simultaneously
at different attention heads within the same layer,
lending itself to favorable computational efficiency
alongside the competence to
preserve fine-grained  details.

\begin{figure}[t]
    \centering
    \begin{tabular}{ccc}
        \includegraphics[width=0.15\textwidth]{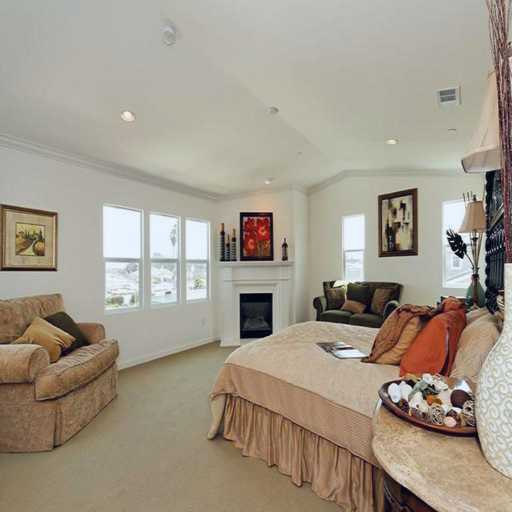}&\hspace{-4mm}
        \includegraphics[width=0.15\textwidth]{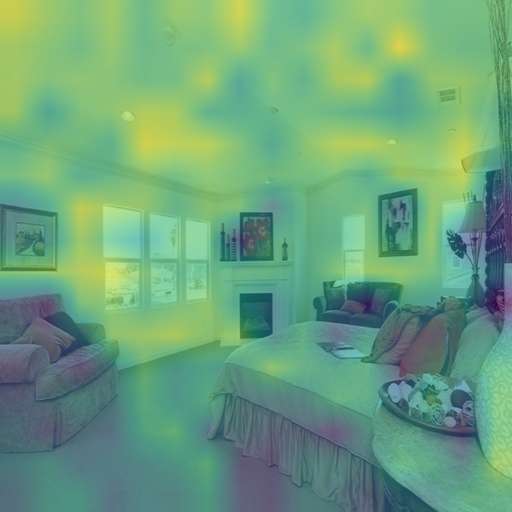}
        &\hspace{-4mm} \includegraphics[width=0.15\textwidth]{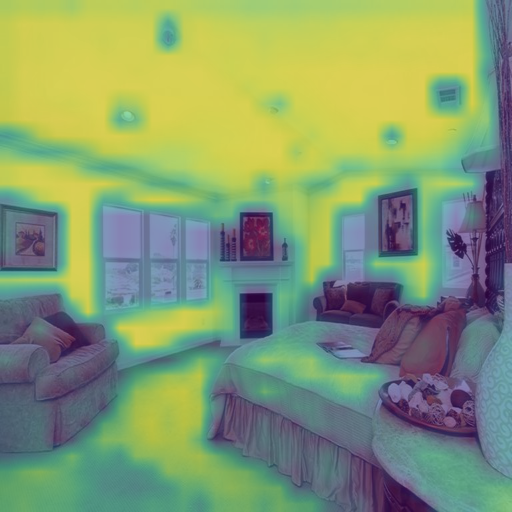}\\
        Image &\hspace{-4mm} PVT &\hspace{-4mm} Ours
    \end{tabular}
    \caption{The attention map of PVT and our model. PVT  attends to  only
    large objects like sofa and bed,
    while our model, by contrast, 
    precisely captures the small objects like lights
    alongside large ones.}
  \label{fig:teaser}
  \vspace{-2mm}
\end{figure}

We show in Figure~\ref{fig:comparison} 
a qualitative comparison between vanilla 
self-attention (from ViT), 
down-sampling aided attention (from PVT), 
and the proposed SSA.
When different attentions 
are applied to features maps of the same size, 
ViT captures fine-grained small objects yet
with an extremely heavy computational cost (Figure~\ref{fig:comparison}(a));
PVT reduces the computation cost 
but its attention is limited only to 
coarse-grained larger objects (Figure~\ref{fig:comparison}(b)).
By contrast, the proposed SSA maintains  
a light computational load yet
simultaneously accounts for hybrid-scale
attentions~(Figure~\ref{fig:comparison}(c)). 
Effectively,
SSA precisely attends to
not only coarse-grained large objects (\eg, sofa)
but also fine-grained small objects (\eg, light and fan),
even some of those located at the corners,
which are unfortunately missed by PVT.
We also show 
visual comparisons of
attention maps in Figure~\ref{fig:teaser},
to highlight the learned 
scale-adaptive attentions of SSA.

The multi-scale attentive mechanism
of SSA is achieved via
splitting 
multiple attention heads into several groups.
Each group   accounts for a dedicated attention  granularity.
For the fine-grained  groups, SSA learns to aggregate few tokens and preserves more local details.
For the remaining coarse-grained head groups, 
SSA learns to aggregate a large amount of tokens and thus reduces computation cost  while preserving the ability of capturing large objects.
The multi-grained groups jointly   learn multi-granularity information,
making the model able to effectively model multi-scale objects.

As depicted in Figure~\ref{fig:sota}, 
we demonstrate the performance of our 
Shunted Transformer model obtained from
stacking multiple SSA-based blocks.
On ImageNet, our Shunted Transformer outperforms 
the state of the art, Focal Transformers~\cite{focal}, 
while halving the model size.
When scaling down to tiny sizes, Shunted Transformer achieves
performance similar to that of DeiT-Small~\cite{deit},
yet with only 50\% parameters. 
For object detection, 
instance segmentation, and semantic segmentation, 
Shunted Transformer consistently outperforms 
Focal Transformer on COCO and ADE20K with a similar model size. 

In sum, our contribution are listed as follows. 
\vspace{-1mm}
\begin{itemize}
\setlength\itemsep{0em}
    \item We propose the Shunted Self-Attention (SSA) which unifies multi-scale feature extractions within one self-attention layer via multi-scale token aggregation. Our SSA adaptively merges tokens on large objects for computation efficiency and preserves the tokens for small objects.
    \item Based on SSA, we build our Shunted Transformer,
    which is able to capture multi-scale objects especially small  and
    remote isolated objects
    efficiently. 
    \item We evaluate our proposed Shunted Transformer on various studies including classification, object detection, and segmentation. Experimental results demonstrate that our Shunted Transformer consistently outperform previous Vision Transformers under similar model sizes.
\end{itemize}

	\section{Related Work}

\subsection{Self-Attention in CNNs}
The receptive field of a convolution layer is usually small and fixed. Although dilated convolution~\cite{yu2015multi} enlarge the receptive filed and deformable convolution allows some offsets~\cite{dai2017deformable} for the kernel, it is hard for them to be   adaptive and flexible to extend to the whole feature maps. Inspired by the self-attention~\cite{vaswani2017attention} layer of transformers pioneered  in the NLP field, some works introduce self-attention or non-local blocks~\cite{wang2018non} to augment convolutional neural networks in the computer vision field. Such attentions always apply in the deep layers, where the size of feature map is small and preprocessed by multiple convolution layers. Therefore, they do not incur too much additional computation cost but bring limited performance improvements. 

\subsection{Vision Transformer}
Vision Transformer (ViT)~\cite{vit} models   directly apply self-attention in very shadow layers to build a convolution-free neural network model. Since the seminal ViT model, many follow-up works are developed to  improve  the model's classification performance~\cite{deit,coadvise} via more complex data augmentation or knowledge distillation.  
Because the computational complexity of self-attention is  quadratic w.r.t. the number of tokens, it is hard for them to directly apply on large number of tokens.  Therefore,  these ViT models usually partition the image into  non-overlapped and large-size patches (tokens). But such partitioning is too coarse and loses much fine-grained information. To preserve fine-grained features, 

these models usually down-sample the feature maps and operate on low-resolution features. This compromise however impedes their deployment in dense-prediction tasks like segmentation and detection. 

\subsection{Efficient ViT Variants}
To make the self-attention attention     applicable on large-size feature maps, recent works develop two solution strategies \cite{swin,focal,pvt,pvtv2,twin}   to reduce the computation cost: (1) split the features maps into   regions and perform local self-attention within the region or (2)  merge tokens to reduce the number of tokens. The representative work  of local attention  is the Swin Transformer~\cite{swin} that  splits feature maps into non-overlap squared regions  and do the self-attention locally. However, to model global dependencies via self-attention,  these local attention  needs to shift the windows over the image  or stack a lot of layers for obtaining a global receptive field. Regarding the strategy of token merging,   PVT (Pyramid Vision Transformer)~\cite{pvt}   designs a spatial-reduction attention to 
merge tokens of key and query. However, PVT and similar models tend to merge too many tokens   in such spatial-reduction. This makes the fine-grained information of small objects mixed with the background and hurts model's performance. Therefore, we propose the shunted self-attention that can simultaneously  preserve  coarse- and fine-grained details while maintaining a global dependency modeling over the image tokens.

\begin{figure*}[t]
    \centering
    \begin{tabular}{c}
        \includegraphics[width=0.98\textwidth]{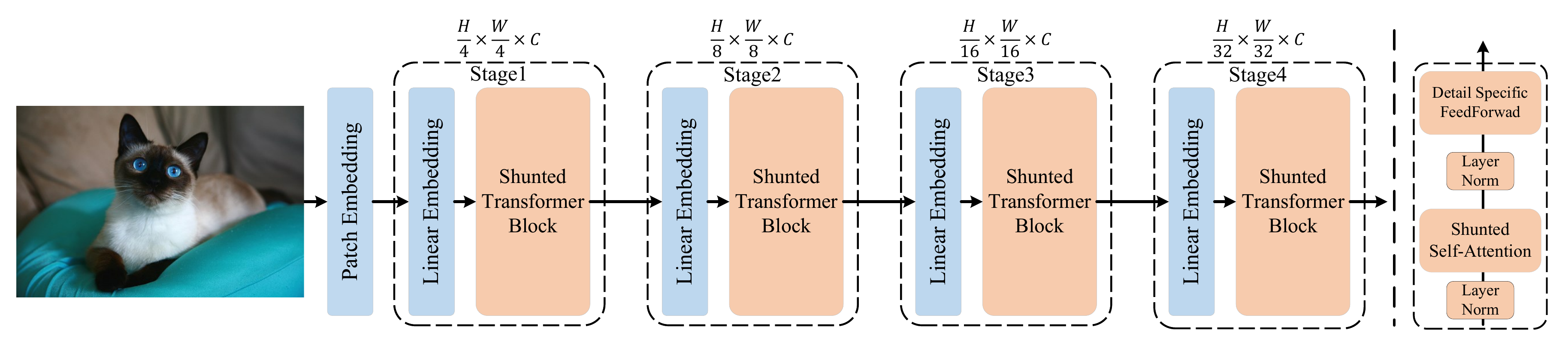}
    \end{tabular}
    \caption{\textbf{Left}: the overall architecture of our Shunted Transformer. \textbf{Right}:   details of our Shunted Self-Attention block.}
  \label{fig:method}
\end{figure*}

	\section{Method}
The overall architecture of our proposed Shunted Transformer is illustrated in Figure \ref{fig:method}. It is built upon the   novel shunted self-attention (SSA) blocks.  There are two main differences between  our SSA blocks and the traditional self-attention blocks   in ViT: 1) SSA introduces a shunted attention mechanism for each  self-attention layer to capture multi-granularity information and better model    objects with different sizes,  especially the small objects;  2) it enhances the capability of extracting local information in the point-wise feed-forward layer by augmenting the cross-token interaction. Besides, our Shunted Transformer deploys a new patch embedding method for obtaining better input feature maps for the first attention block.  In the following, we elaborate on these novelties one by one. 
    
\subsection{Shunted Transformer Block}
\begin{figure}[t]
    \centering
    \begin{tabular}{c}
        \hspace{-3mm}\includegraphics[width=0.4\textwidth]{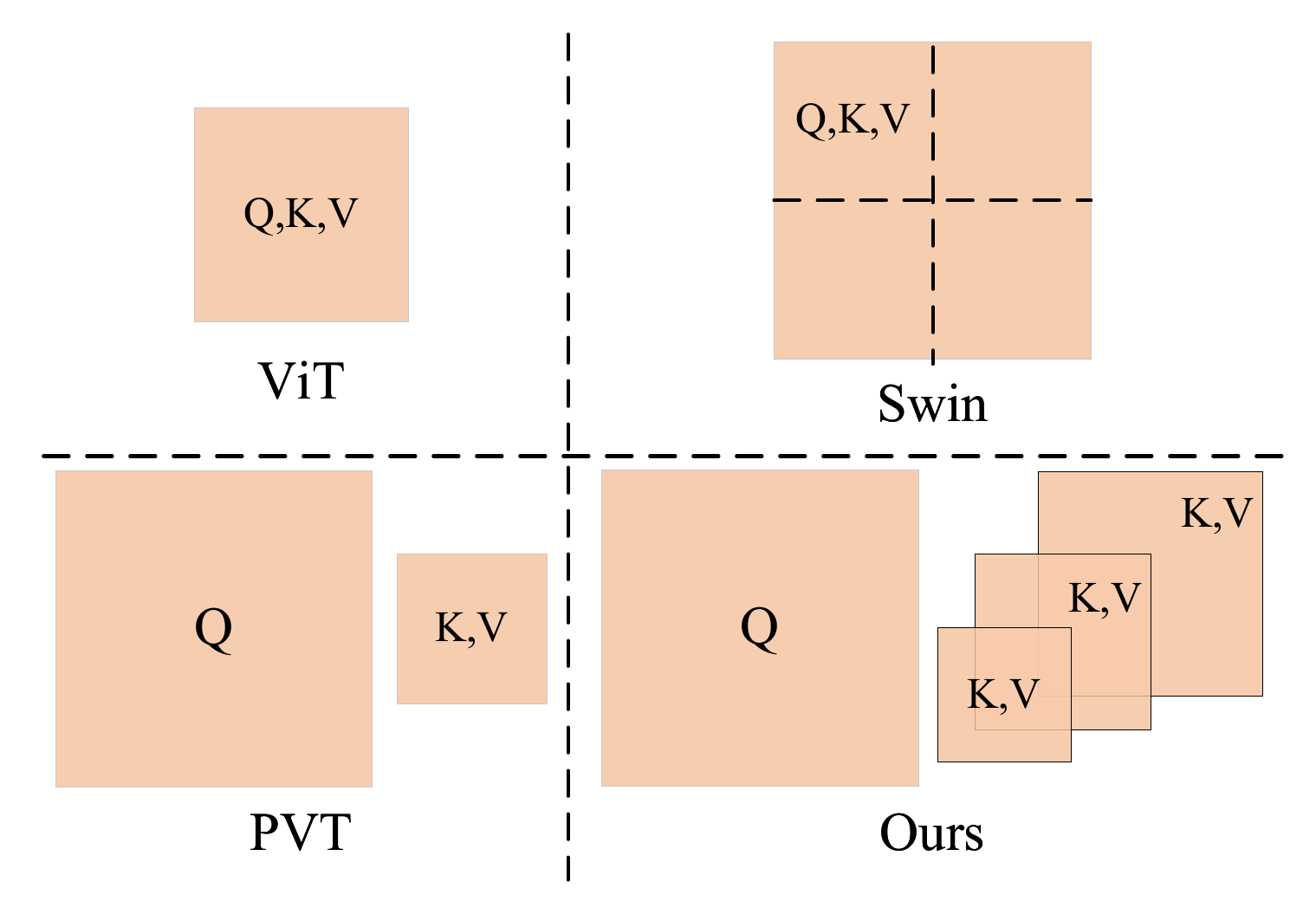}
    \end{tabular}
    \caption{Comparing our shunted self-attention with self attention in ViT, Swin, PVT. ViT applies self-attention globally on small-size feature maps. Swin Transformer applies local self-attention on large-size feature maps within small  regions. PVT fuses the key and value with a large stride. Differently, our shunted self-attention conducts multi-scale token aggregation for obtaining key and value of various sizes.}
    \vspace{-2mm}
  \label{fig:attention}
\end{figure}
In the $i$-th stage of the proposed Shunted Transformer, there are $L_i$ transformer blocks. Each transformer block contains a self-attention layer and a feed-forward layer. To reduce the  computation cost when processing high-resolution feature maps, PVT~\cite{pvt} introduces spatial-reduction attention (SRA) to replace the original multi-head self-attention (MSA). However, SRA tends to merage too many tokens    within one self-attention layer and only provides token features at a single   scale. These limitations  impede the capability of the models in capturing multi-scale objects especially the small-size ones. Therefore, we introduce our shunted self-attention with learning multi-granularity within one self-attention layer in parallel.

\subsubsection{Shunted Self-Attention}
The input sequence $F \in \mathbb{R}^{h\times w \times c}$ are projected into query ($Q$), key ($K$) and value ($V$) tensors at first. Then the multi-head self-attention adopts  $H$  independent attention heads to compute self-attention in parallel. 
To reduce the computation cost, we follow the PVT \cite{pvt} and reduce the length of $K$ and $V$ instead of splitting $\{Q, K, V\}$ into regions as in Swin Transformer \cite{swin}. 

As show in Figure \ref{fig:attention}, our SSA is different from the SRA of  PVT in that the length of $K$, $V$ is not identical across the attention heads of the same self-attention layer. Instead, the length varies in different heads for capturing different granularity information. This gives  the multi-scale token aggregation (MTA). Specifically,  the keys $K$ and values $V$ are down-sampled to different sizes for different heads indexed by $i$:
\begin{equation}
\begin{aligned}
    Q_i &= XW^Q_i, \\
    K_i, V_i &= MTA(X, r_i)W^K_i,~MTA(X, r_i)W^V_i, \\
    V_i &= V_i + LE(V_i).
\end{aligned}
\end{equation}
Here the $\mathrm{MAT}(\cdot; r_i)$ is the multi-scale token aggregation layer in the $i$-th head with the down-sampling rate of $r_i$. In practice, we take a convolution layer with kernel size and stride of $r_i$ to implement the down-sampling. $W^Q_i, W^K_i, W^V_i$ are the parameters of the linear projection in the $i$-th head. There are variant $r_i$ in one layer across the attention heads. Therefore, the key and value can capture different scales    in a self-attention.  $LE(\cdot)$ is the local enhancing component of MTA for value $V$ by a depth-wise convolution.    Comparing with the spatial-reduction~\cite{pvt}, more fine-grained and low-level details are preserved. 

Then the shunted self-attention is calculated by:
\begin{equation}
    h_i = \mathrm{Softmax}\left(\frac{Q_iK_i^\mathsf{T}}{\sqrt{d_h}}\right) V_i 
\end{equation}
where $d_h$ is the dimension. Thanks to multi-scale key and value, our shunted self-attention is more powerful in capturing multi-scale objects. The computation cost reduction may depend on the value of $r$, therefore, we can well define the model and $r$ to trade-off the computation cost and model performance. When $r$ grows large, more tokens in $K,V$ are merged and the length of $K,V$ is shorter, therefore, the computation cost is low but it still preserve the ability of capturing large objects. In contrast, when $r$ becomes small, more details are preserved but brings more computation cost. Integrating various $r$ in one self-attention layer enables  it  to capture multi-granularity features.

\begin{table*}[t]
\centering
\small
\begin{tabular}{c|c|c|c|c|c}
\toprule
                        & Output Size              & Layer Name  & Shunted-Tiny                & Shunted-Small & Shunted-Base \\
\midrule
\multirow{2}{*}{Stage1} & \multirow{2}{*}{56x56}   & \multirow{2}{*}{\makecell[c]{Transformer \\ Block}} &
\makecell[c]{$r_i$= 
$\begin{cases}
4& \text{i  \textless  $\frac{head}{2}$}\\
8& \text{i $\geq \frac{head}{2}$}
\end{cases}$}&
\makecell[c]{$r_i$= 
$\begin{cases}
4& \text{i  \textless  $\frac{head}{2}$}\\
8& \text{i $\geq \frac{head}{2}$}
\end{cases}$}&
\makecell[c]{$r_i$= 
$\begin{cases}
4& \text{i  \textless  $\frac{head}{2}$}\\
8& \text{i $\geq \frac{head}{2}$}
\end{cases}$}           \\
&&& $C_1$=64, $head$=2, $N_1$=1         &   $C_1$=64, $head$=2, $N_1$=2       &     $C_1$=64, $head$=2, $N_1$=3      \\
\midrule
\multirow{2}{*}{Stage2} & \multirow{2}{*}{28x28}   & \multirow{2}{*}{\makecell[c]{Transformer \\ Block}} &    
\makecell[c]{$r_i$= 
$\begin{cases}
2& \text{i  \textless  $\frac{head}{2}$}\\
4& \text{i $\geq \frac{head}{2}$}
\end{cases}$}&\makecell[c]{$r_i$= 
$\begin{cases}
2& \text{i  \textless  $\frac{head}{2}$}\\
4& \text{i $\geq \frac{head}{2}$}
\end{cases}$}          &\makecell[c]{$r_i$= 
$\begin{cases}
2& \text{i  \textless  $\frac{head}{2}$}\\
4& \text{i $\geq \frac{head}{2}$}
\end{cases}$}           \\
 & &  &$C_2$=128, $head$=4, $N_1$=2        &   $C_2$=128, $head$=4, $N_1$=4       &     $C_2$=128, $head$=4, $N_1$=4      \\
\midrule
\multirow{2}{*}{Stage3} & \multirow{2}{*}{14x14}   & \multirow{2}{*}{\makecell[c]{Transformer \\ Block}} &
\makecell[c]{$r_i$= 
$\begin{cases}
1& \text{i  \textless  $\frac{head}{2}$}\\
2& \text{i $\geq \frac{head}{2}$}
\end{cases}$}&\makecell[c]{$r_i$= 
$\begin{cases}
1& \text{i  \textless  $\frac{head}{2}$}\\
2& \text{i $\geq \frac{head}{2}$}
\end{cases}$}          &\makecell[c]{$r_i$= 
$\begin{cases}
1& \text{i  \textless  $\frac{head}{2}$}\\
2& \text{i $\geq \frac{head}{2}$}
\end{cases}$}           \\
&&&$C_3$=256, $head$=8, $N_1$=4    &  $C_3$=256, $head$=8, $N_1$=12        &  $C_3$=256, $head$=8, $N_1$=24         \\
\midrule
\multirow{2}{*}{Stage4} & \multirow{2}{*}{7x7}     & \multirow{2}{*}{\makecell[c]{Transformer \\ Block}} &
$r$= 1& $r$= 1        &  $r$= 1      \\
&&        &$C_4$=512, $head$=16, $N_1$=1          &   $C_4$=512, $head$=16, $N_1$=1       &    $C_4$=512, $head$=16, $N_1$=2   \\   
\bottomrule
\end{tabular}
\caption{Model variants for our Shunted Transformer. $C$ and $N$ represent the dimension and number of blocks. $head$ indicates the number of heads. }
\label{variant}
\end{table*}

\subsubsection{Detail-specific   Feedforward Layers}
\begin{figure}[t]
    \centering
    \begin{tabular}{c}
        \hspace{-3mm}\includegraphics[width=0.48\textwidth]{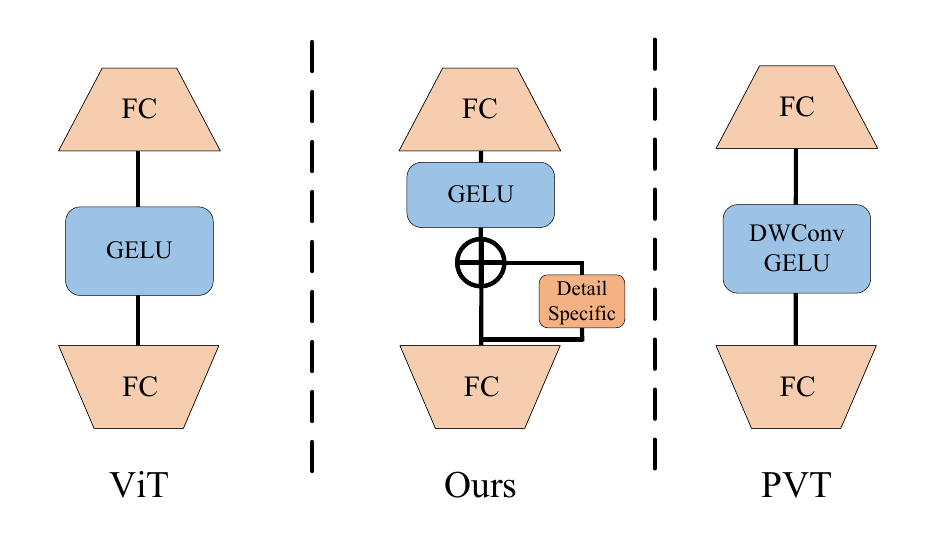}
    \end{tabular}
    \caption{Comparing the feed-forward layer in ViT (left), PVT (right), and our detail-specific feedfoward layer. We complement fine-grained cross-token details in the feed-forward layer.}
  \label{fig:mlp}
\end{figure}

In the traditional feed forward layer, the fully connected layer are point-wise and no cross token information can be learnt. Here, we aim at complementing local information by specifying the details in the feedforward layer. As shown in Figure \ref{fig:mlp}, we   complement the local details in the feed forward layer by adding our data specific layer between the two fully connected layer in the feed forward layer:
\begin{equation}
\begin{split}
    x' &= FC(x; \theta_1),\\
    x'' &= FC(\sigma(x'+DS(x';\theta));\theta_2),
\end{split}
\end{equation}
where $DS(\cdot;\theta)$ is the detail specific layer with parameters $\theta$, implemented by a depth-wise convolution in practice.

\subsection{Patch Embedding}
Transformer is firstly designed for handling  sequential data. How to map the image to sequence is important for the model's performance. ViT directly splits the input image into $16\times 16$ non-overlap patches. A recent study~\cite{wang2021scaled} finds using convolution in the patch embedding provides a higher-quality token sequence   and helps transformer ``see better" than  a conventional large-stride non-overlapping patch embedding. Therefore, some works~\cite{pvt,swin} conduct  overlapped patch embedding   like using a $7\times 7$ convolution. 

In our model, we take different convolution layers with overlapping based  on the model size. We take a $7\times 7$ convolution layer with stride of 2 and zero padding as the first layer in the patch embedding, and add extra $3\times 3$ convolution layer with stride of 1 depending on the model size. Finally, a non-overlapping projection layer with stride of 2 to generate the input sequence with size of $\frac{H}{4} \times \frac{W}{4}$.

\subsection{Architecture Details and Variants}

Given an input image with size of $H\times W \times 3$, we adopt the above patch embedding scheme for obtaining   more informative token sequence 
with the length of $\frac{H}{4} \times \frac{W}{4}$ and the token dimension of $C$. 
Following previous designs~\cite{swin,focal,pvt,twin}, there are four stages in our model and each stage contain several Shunted Transformer blocks. In each stage, each block outputs the feature maps of the same  size. We take a convolution layer with stride 2 (Linear embedding) to connect different stages and the size of the feature maps will be halved before feeding into  to the next stage, but the dimension will be doubled. Therefore, we have four feature maps ${F_1, F_2, F_3, F_4}$ of the output of each stage and the size of $F_i$ is $\frac{H}{2^{i+1}} \times \frac{W}{2^{i+1}} \times (C\times 2^{i-1})$. 

We propose three kinds of different configurations of our model for fair comparison under similar parameters and computation costs. As show in Table \ref{variant}, $head$ and the $N_i$ indicate the number of heads in one block and the number of blocks in one stage. The variants only comes from the number of layers in different stage. Specifically, the number of head in each block is set 2,4,8,16. The convolution in the patch embedding range from 1 to 3. 

	\section{Experiments}

\begin{table}[t]
\centering
\small
\begin{tabular}{l|c|c|c}
\toprule
\multirow{2}{*}{Model} & ~Params~ & ~~FLOPs~~ & ~~Top1~~ \\
                       & (M)     & (G) &     (\%) \\
\midrule
ResNet-18~\cite{resnet}    &   11.7         &     1.8      &    69.8      \\
RegNetY-1.6G~\cite{regnet} &     11.2       &   1.6        &       78.0   \\
DeiT-T~\cite{deit}       &     5.7       &     1.3      &     72.2    \\
PVT-T~\cite{pvt}       &       13.2     &   1.9        &   75.1      \\
PVTv2-b1~\cite{pvtv2}       &    13.1 &    2.1      &    78.7      \\
\rowcolor{mygray}
Shunted-T  &       11.5     &   2.1        &      \textbf{79.8}   \\
\midrule
ResNet-50~\cite{resnet}    &       25.0     &     4.1      &    76.2      \\
RegNetY-4G~\cite{regnet} &       20.6     &  4.0         &     79.4     \\
EfficientNet-B4*~\cite{tan2019efficientnet} &       19    &  4.2        &     82.9     \\
DeiT-S~\cite{deit}         &      22.1      & 4.6          & 79.9        \\
T2T-14~\cite{yuan2021tokens} & 22.0 &5.2 &81.5 \\
DeepViT-S~\cite{zhou2021deepvit} & 27.0 &6.2 &82.3 \\
ViL-S~\cite{zhang2021multi} & 24.6 &4.9 &82.0 \\
TNT-S~\cite{han2021transformer} &23.8  &5.2 & 81.3\\
CViT-15~\cite{chen2021crossvit}& 27.4 & 5.6&81.5 \\
PVT-S~\cite{pvt}       &      24.5      &     3.8      &   79.8      \\
Swin-T~\cite{swin} &      28.3      &   4.5        &    81.2     \\
Twin-S~\cite{twin} & 24 &2.9 &81.7 \\
Focal-T~\cite{focal}  &        29.1    &    4.9       &    82.2     \\
PVTv2-b2~\cite{pvtv2}      &    25.4      &      4.0       &    82.0      \\
\rowcolor{mygray}
Shunted-S &        22.4    &      4.9     &     \textbf{82.9}    \\
\midrule
ResNet-101~\cite{resnet}       &     45.0     &    7.9     &  77.4     \\
ViT-B~\cite{vit} &     86.6     &    17.6     &   77.9     \\
DeiT-B~\cite{deit}   &     86.6     &     17.5     &  81.8    \\
Swin-S~\cite{swin} &      49.6     &   8.7   &   83.1     \\
Swin-B~\cite{swin} &      87.8    &   15.4   &   83.4     \\
PVT-M~\cite{pvt}  &     44.2     &     6.7    &   81.2   \\
PVT-L~\cite{pvt}  &     61.4    &    9.8  &   81.7   \\
Focal-S~\cite{focal} &      51.1     &     9.1    &  83.5    \\
Focal-B~\cite{focal} &      89.8    &    16.0   &  83.8    \\
\rowcolor{mygray}
Shunted-B       &     39.6   &    8.1      &   \textbf{84.0}      \\
\bottomrule
\end{tabular}
\vspace{-1mm}
\caption{Comparison of different backbones on ImageNet-1K classification. Except EfficientNet (EffNet-B4*), all models are trained and evaluated on the input size of $224\times 224$.}
\label{classification}
\end{table}

To evaluate the effectiveness of our Shunted Transformer, we apply our model on ImageNet-1K~\cite{deng2009imagenet} classification, COCO~\cite{coco} object detection and instance segmentation, ADE20K~\cite{ade20k} semantic segmentation. Besides, we evaluate effects of different components of our model via ablation studies.

\subsection{Image Classification on ImageNet-1K}
We first evaluate our   model and compare it with recent SOTA CNN and transformer based models on  ImageNet-1K. For fair comparison, we follow the same training strategies of DeiT~\cite{deit} and PVT~\cite{pvt}. Specifically, we take AdamW as the optimizer with the weight decay of 0.05. The whole training epochs are 300 with the input size of $224\times 224$, and the batch size is 1024. The learning rate is set to 1$\times 10^{-3}$ following cosine learning rate decay. The data augmentations and regularization methods follow DeiT \cite{deit} including random cropping, random flipping, label smoothing~\cite{szegedy2016rethinking}, Mixup~\cite{mixup}, CutMix~\cite{cutmix} and random erasing~\cite{zhong2020random}. 

As shown in Table \ref{classification}, by comparing   with other CNN backbones under similar parameters and computation cost,  our model is  the first transformer based model that achieves   comparable results with EfficientNet which uses   much larger input resolution. Notably, although RegNet and EfficientNet come  from neural architecture search, our manually designed Transformer still outperform them.

We then compare  our model with Transformer backbones. Our tiny model achieves similar performance with Transformer baseline (DeiT-S) but only requires half of parameters (22M$\rightarrow $11M) and computation cost (4.6G$\rightarrow $2.1G FLOPs). When our model size grows similar to DeiT-S, it outperforms by  3\%. Comparing with the very recent SOTA models like Swin and Twin, our model consistently outperform them. Specifically, our small-size model outperforms the existing state-of-the-art, Focal Transformer Tiny by 0.7\%, while reducing the model size by 20\%. When model size grows large, our base model achieve state-of-the-art performance with only half of parameters and computation cost comparing with Focal Transformer.

\subsection{Object Detection and Instance Segmentation}
\begin{table*}[t]
\small
{
\begin{tabular}{l|c|cccccc|cccccc}
\toprule
\multirow{2}{*}{Backbone} & Params &  \multicolumn{6}{c|}{Mask R-CNN 1$\times$ schedule} & \multicolumn{6}{c}{Mask R-CNN 3$\times$ schedule + MS} \\

                          &   (M)       &   $AP^b$    &  $AP^b_{50}$     &  $AP^b_{75}$     &   $AP^m$    &   $AP^m_{50}$    &  $AP^m_{75}$     &    $AP^b$    &  $AP^b_{50}$     &  $AP^b_{75}$     &   $AP^m$    &   $AP^m_{50}$    &  $AP^m_{75}$     \\
\midrule
Res50~\cite{resnet} &   44        &   38.0& 58.6 &41.4 &34.4& 55.1& 36.7 &41.0& 61.7& 44.9& 37.1& 58.4& 40.1 \\
PVT-S~\cite{pvt}    &    44       & 40.4& 62.9& 43.8& 37.8& 60.1& 40.3& 43.0& 65.3 &46.9& 39.9& 62.5& 42.8 \\
Swin-T~\cite{swin} & 48 & 42.2 &64.6& 46.2 &39.1 &61.6& 42.0 &46.0& 68.2 &50.2 &41.6& 65.1 &44.8\\
TwinP-S~\cite{twin} & 44 & 42.9 &65.8& 47.1 &40.0 &62.7& 42.9 &46.8& 69.3 &51.8 &42.6& 66.3 &46.0\\
Twin-S~\cite{twin} & 44&  43.4 &66.0& 47.3 &40.3 &63.2& 43.4 &46.8& 69.2 &51.2 &42.6& 66.3 &45.8\\
Focal-T~\cite{focal} & 49 & 44.8 &67.7& 49.2& 41.0& 64.7& 44.2 & 47.2 &69.4 &51.9& 42.7& 66.5& 45.9 \\
PVTv2-b2*~\cite{pvtv2} & 45& 45.3& 67.1& 49.6& 41.2 &64.2 &44.4&-&- &- &- & -&- \\
\rowcolor{mygray}
Shunted-S & \textbf{42}& \textbf{47.1}& \textbf{68.8}& \textbf{52.1} &\textbf{42.5}& \textbf{65.8} &\textbf{45.7} &\textbf{49.1} &\textbf{70.6} &\textbf{53.8} &\textbf{43.9} &\textbf{67.8} &\textbf{47.5}\\
\midrule
Res101~\cite{resnet}  &   63      &  40.4& 61.1 &44.2 &36.4& 57.7& 38.8 &42.8& 63.2& 47.1& 38.5& 60.1& 41.3 \\
PVT-M~\cite{pvt}    &    64        & 42.0& 64.4& 45.6& 39.0& 61.6& 42.1& 44.2& 66.0 &48.2& 40.5& 63.1& 43.5 \\
Swin-S~\cite{swin} & 69 & 44.8 &66.6& 48.9 &40.9 &63.4& 44.2 &48.5& 70.2 &53.5 &43.3& 67.3 &46.6\\
Swin-B~\cite{swin} & 107 & 46.9 &-& - &42.3 &-& - &48.5& 69.8 &53.2 &43.4& 66.8 &46.9\\
TwinP-B~\cite{twin} & 64 & 44.6 &66.7& 48.9 &40.9 &63.8& 44.2 &47.9& 70.1 &52.5&43.2& 67.2 &46.3\\
Twin-B~\cite{twin} & 76& 45.2 &67.6& 49.3 &41.5 &64.5& 44.8 &48.0& 69.5 &52.7 &43.0&66.8 &46.6\\
Focal-S~\cite{focal} & 71& 47.4 &\textbf{69.8} & 51.9& 42.8& 66.6& 46.1 & 48.8 &70.5 &53.6& 43.8& 67.7& 47.2 \\
\rowcolor{mygray}
Shunted-B & \textbf{59} & \textbf{48.0} &\textbf{69.8} &\textbf{53.3} &\textbf{43.2} & \textbf{66.9}& \textbf{46.8} & \textbf{50.1} &\textbf{70.9} &\textbf{54.1} & \textbf{45.2}& \textbf{68.0}& \textbf{48.0} \\
\bottomrule
\end{tabular}}
\vspace{-3mm}
\caption{Object detection and instance segmentation with Mask R-CNN on COCO. Only 3$\times$ schedule has the multi-scale training. All backbone are pretrained on ImageNet-1K. * indicate that methods have not been peer reviewed.}
\label{maskrcnn}
\end{table*}

\begin{table*}[t]
\small
{
\begin{tabular}{l|c|cccccc|cccccc}
\toprule
\multirow{2}{*}{Backbone} & Params &  \multicolumn{6}{c|}{RetinaNet 1$\times$ schedule} & \multicolumn{6}{c}{RetinaNet 3$\times$ schedule + MS} \\

                          &   (M)       &   $AP^b$    &  $AP^b_{50}$     &  $AP^b_{75}$     &   $AP_S$~    &   $AP_M$~    &  $AP_L$~     &    $AP^b$    &  $AP^b_{50}$     &  $AP^b_{75}$     &   $AP_S$~    &   $AP_M$~    &  $AP_L$     \\
\midrule
Res50~\cite{resnet} &   37.7        &   36.3& 55.3 &38.6 &19.3& 40.0& 48.8 &39.0& 58.4& 41.8& 22.4& 42.8& 51.6 \\
PVT-S~\cite{pvt}    &    34.2       & 40.4& 61.3& 43.0& 25.0& 42.9& 55.7& 42.2& 62.7 &45.0& 26.2& 45.2& 57.2 \\
ViL-S~\cite{pvt}    &    35.7       & 41.6& 62.5& 44.1& 24.9& 44.6& 56.2& 42.9& 63.8 &45.6& 27.8& 46.4& 56.3 \\
Swin-T~\cite{swin} & 38.5  & 42.0 &63.0& 44.7 &26.6 &45.8& 55.7 &45.0& 65.9 &48.4 &29.7& 48.9 &58.1\\
Focal-T~\cite{focal} & 39.4 & 43.7 &65.2& 46.7& 28.6& 47.4& 56.9 & 45.5 &66.3 &48.8& \textbf{31.2}& 49.2& 58.7 \\
PVTv2-b2*~\cite{pvtv2} & 35.1& 44.6& 65.6& 47.6& 27.4&48.8 &58.6&-&- &- &- & -&- \\
\rowcolor{mygray}
Shunted-S & \textbf{32.1}& \textbf{45.4}&\textbf{65.9} &\textbf{49.2} &\textbf{28.7}&\textbf{49.3}&\textbf{60.0}& \textbf{46.4}& \textbf{66.7}&\textbf{50.4} &31.0 &\textbf{51.0} &\textbf{60.8}\\
\bottomrule
\end{tabular}}
\vspace{-3mm}
\caption{Object detection with RetinaNet on COCO. Only 3$\times$ schedule has the multi-scale training. All backbone are pretrained on ImageNet-1K. * indicate that methods have not been peer reviewed.}
\label{retina}
\vspace{-4mm}
\end{table*}

We evaluate the models for object detection and instance segmentation on COCO 2017~\cite{coco}.  We take our proposed Shunted Transformer as backbone and plug it into Mask R-CNN. We compare it  with other SOTA backbones including ResNet, Swin Transformer, Pyramid Vision Transformer, Twin and Focal Transformer. We follow the same settings of Swin: pretraining on ImageNet-1K and fine-tuning on COCO. In the fine-tuning stage, we take two training schedules: 1$\times$ with 12 epochs and 3$\times$ with 36 epochs. In 1$\times$ schedule, the shorter side of the input image will be resize to 800 while keeping the longer side no more than 1333. In 3$\times$ schedule, we take multi-scale training strategy of resizing the shorter size between 480 to 800. We take AdamW with weight decay of 0.05 as the optimizer. The batch size is 16 and initial learning rate is $10^{-4}$.

In Table \ref{maskrcnn}, we take Mask-RCNN for object detection and report the bbox mAP ($AP^b$) of different CNN and Transformer   backbones. Under comparable parameters, our model outperforms previous SOTA with a significant gap. For object detection, with 1$\times$ schedule, our tiny model achieves 9.1 points improvements over ResNet-50, and 2.3 points over   Focal Transformer with only 85\% model size. Moreover, with 3$\times$ schedule and multi-scale training, our backbone still concisely outperforms CNN backbones over 7.7 and Transformer backbone over 1.6 points on average. We find similar results in instance segmentation. We report the mask mAP ($AP^m$) in Table \ref{maskrcnn}. Our model achieves 8.1 points higher than ResNet-50 and 1.5 points higher than Focal Transformer in 1$\times$ schedule and in 3$\times$ schedule. Our model  achieves these superior performances at  a smaller model size, clearly demonstrating the benefits of its shunted attention  for learning multi-granularity tokens and effectiveness in handling   presence of multi-scale visual objects.

We also report the results of RetinaNet in the Table \ref{retina}. With the least parameters, our model outperforms  all the previous ones in both 1$\times$ and 3$\times$ schedule. Comparing with       PVT, our model brings improvements on all small, medium and large size objects which shows the strong power of capturing multi-scale objects in our shunted self-Attention.

\subsection{Semantic Segmentation on ADE20K}
\begin{table*}[t]
\centering
{
\small
\begin{tabular}{l|ccc|cccc}
\toprule
\multirow{2}{*}{Backbone} & \multicolumn{3}{c|}{Semantic FPN 80k} & \multicolumn{4}{c}{Upernet 160K}       \\
                          & Param (M)    & FLOPs (G)    & mIOU (\%)   & Param (M) & FLOPs (G) & mIOU (\%) & MS mIOU (\%)\\
\midrule
ResNet-50~\cite{resnet} &       28.5       &    183          &  36.7 &  -     &       -    &   -   &   - \\
Swin-T~\cite{swin}   &    31.9     &     182    &    41.5  &    59.9    &      945     & 44.5     &  45.8  \\
PVT-S~\cite{pvt} &28.2      &    116    & 39.8   &     -    &   - & -  & -  \\
TwinsP-S~\cite{twin}       &     28.4     &    162   & 44.3       & 54.6      &   919    &   46.2 &47.5    \\
Twin-S~\cite{twin}       &   28.3   &  144         &43.2&54.4           & 901          &46.2      &47.1   \\
Focal-T~\cite{focal}    &    -          &      -   &   -     &62           &998           &45.8      &   47.0 \\
\rowcolor{mygray}
Shunted-S    &       26.1       &     183    &  48.2     &52           &940           &48.9      &   49.9 \\
\bottomrule 
\end{tabular}}
\vspace{-3mm}
\caption{Comparison of the segmentation performance of different backbones in Semantic FPN and UpperNet framework on ADE20K.  }
\label{seg}
\end{table*}

We evaluate the performance of our model for semantic segmentation on the ADE20K~\cite{ade20k} dataset. There are 20,210 images for training, 2,000 images for validation and 3,352 images for testing with 150 fine-grained semantic categories. We report the mIOU with and without multi-scale testing. We take UperNet and Semantic FPN as the main frameworks and take different architectures as backbones. We follow the defaults settings of Focal Transformer and mmsegmentation~\cite{mmseg2020}. For UpperNet, we take AdamW with weight decay of 0.01 as the optimizer for 160K iterations. The learning rate is $6\times 10^{-5}$ with 1500 iteration warmup at the begining of training and linear learning rate decay. The augmentations include random flipping, random scaling and random photo-metric distortion. The input size is $512\times 512$ in training, and single scale and multi-scale (MS) test. For SemanticFPN, we take AdamW with weight decay of 0.0001 as the optimizer and the learning rate is also 0.0001 for 80K iterations.

The results are reported in Table \ref{seg}. Our Shunted Transformer outperforms previous state-of-the-art with a large margin and less parameters for all the frameworks. Specifically, when using semantic FPN, our model outperforms the Swin Transformer by 6.7 mIOU,  with   20\% model size reduction. When the framework is UpperNet, our Shunted Transformer is 3.1\% and 2.9\% higher than focal transformer. The results of segmentation the shows the superiority of our Shunted Transformer. 

We also take SegFormer~\cite{xie2021segformer} as the framework and compare our backbone with the MiT in the SegFormer. The results are reported in Table \ref{segformer}. With less parameters, our method achieve 1.8 mIoU improvements over SegFormer.

\begin{table}[h!]
\centering
{
\small
\begin{tabular}{l|c|c|c}
\toprule
Backbone& Params(M)  & FLOPs (G)  & mIoU \\
\midrule
MiT-B2    &   27.5       &   62.4    &    46.5      \\
\rowcolor{mygray}
Ours       &    25.1      &  70.3    &    48.3  \\
\bottomrule
\end{tabular}}
\vspace{-1mm}
\caption{Comparison of different backbone in Segformer framework on Ade20K.}
\label{segformer}
\vspace{-2mm}
\end{table}

\subsection{Ablation Studies}

\paragraph{Patch Embedding} 
Many recent works\cite{head1,head2,head3} study the function of the image to token mapping, \ie the patch embedding head. They find well-designed head provide better input sequence for the transformer models. We evaluate the impact of our patch embedding with non-overlap head in ViT, overlap head in Swin and PVT. The results are shown in Table \ref{head}. With more complex head like overlapping or our patch embedding, the computation cost and model size only slightly grow, but the performance improves relatively significant. Specifically, with limited additional parameters, from the traditional non-overlap head or the overlap head to patch embedding, the model achieves 1.4\% and 0.3\% performance gain respectively. 

\begin{table}[h!]
\centering
{
\small
\begin{tabular}{l|c|c|c}
\toprule
Patch Embedding& Params (M)  & FLOPs (G)  & Top-1 (\%)  \\
\midrule
Non-Overlap    &   22.3       &   4.4    &    81.5      \\
Overlap &     22.4       &  4.5       &       82.6  \\
\rowcolor{mygray}
Ours       &     22.4      &    4.9     &     82.9   \\
\bottomrule
\end{tabular}
}\vspace{-1mm}
\caption{Top-1 accuracy on ImageNet of different patch embedding heads. Our patch embedding requires slightly more computation cost, but the performance improvement is significant.}
\label{head}
\vspace{-4mm}
\end{table}

\vspace{-3mm}
\paragraph{Token Aggregation Function}
We propose a new  token aggregation function to merge tokens for multi-scale objects and keeping the global and local information simultaneously. From Table \ref{aggregation}, our novel token aggregation function has similar computation with the convolutional spatial-reduction but gain more improvements.

\begin{table}[h!]
\centering
{
\small
\begin{tabular}{l|c|c|c}
\toprule
Aggregation & Params(M)  & FLOPs (G)  & Top-1 (\%)  \\
\midrule
Linear    &   18.5       &   4.5    &    82.1      \\
Convolution &     22.4       &  4.9      &       82.6  \\
\rowcolor{mygray}
Ours       &     22.4      &    4.9    &     82.9   \\
\bottomrule
\end{tabular}}\vspace{-1mm}
\caption{Top-1 accuracy on ImageNet of different token aggregation functions.}
\label{aggregation}
\vspace{-2mm}
\end{table}

\vspace{-5mm}
\paragraph{Detail-specific Feed-Forward}
In the Feed-Forward layer~\cite{deit}, all the operation is point-wise and no cross token operations exists, therefore, complement the cross token and local information will significantly improves the learning ability of the feed-forward layer. In   Table \ref{mlp}, we compare our new detail-specific feed-forward layer, traditional feed-forward layer~\cite{deit} and convolutional feed-forward layer~\cite{pvtv2} in ViT and our model. The detail-specific feed-Forward consistently brings performance gain over the traditional feedforward layer which indicate the utility of complementing the local details in the feedforward layer.

\begin{table}[h!]
\centering
{
\small
\begin{tabular}{l|c|c}
\toprule
Layers&  Backbone  & Top-1 (\%)  \\
\midrule
Feedforward    &   ViT       &   79.8     \\
Conv-Feedforward &     ViT       &       80.5  \\
\rowcolor{mygray}
Detail Specific FeedForward       &     ViT     &  80.7  \\
\midrule
Feedforward    &   Shunted       &   82.6     \\
Conv-Feedforward &     Shunted       &       82.7  \\
\rowcolor{mygray}
Detail Specific FeedForward       &     Shunted     &  82.9 \\
\bottomrule
\end{tabular}}\vspace{-1mm}
\caption{With similar parameter numbers and FLOPs,  Detail-specific feedForward layers provide higher top-1 accuracy on ImageNet than the traditional feedforward ones.}
\label{mlp}
\vspace{-2mm}
\end{table}

	\vspace{-3mm}
	\section{Conclusion}
	In this paper, we present a novel Shunted Self-Attention~(SSA)
	scheme to explicitly account for multi-scale features.
	In contrast to prior works that 
	focus on only static feature maps in one attention layer,
	we maintain various-scale feature maps
	that attend to multi-scale objects within one self-attention layer. 
	Extensive experiments show the effectiveness of our model as a backbone 
	for various downstream tasks. 
	Specifically, the proposed model
	outperforms prior Transformers,
	and achieves 
	state-of-the-art results on classification, 
	detection, and segmentation tasks.
	
	\section*{Acknowledgement}
	This work is supported by NUS Faculty Research Committee Grant (WBS: A-0009440-00-00) and NRF Centre for Advanced Robotics Technology Innovation (CARTIN). 
	
	{\small
		\bibliographystyle{ieee_fullname}
		\bibliography{egbib}
	}
	
\end{document}